\documentclass{article} % For LaTeX2e
\usepackage{times}
\usepackage{iclr/iclr_conference}
\usepackage{comment}
\usepackage{amsmath,amsfonts,bm}

\def\eqref#1{equation~\ref{#1}}
\def\1{\bm{1}}

\DeclareMathAlphabet{\mathsfit}{\encodingdefault}{\sfdefault}{m}{sl}
\SetMathAlphabet{\mathsfit}{bold}{\encodingdefault}{\sfdefault}{bx}{n}

\usepackage{floatrow}
\newfloatcommand{capbtabbox}{table}[][\FBwidth]
\usepackage{blindtext}
\usepackage[colorlinks=true]{hyperref}       % hyperlinks
\usepackage{url}
\usepackage{booktabs}       % professional-quality tables
\usepackage{nicefrac}       % compact symbols for 1/2, etc.
\usepackage{amsmath,amsfonts,amssymb,multirow,amsthm}
\usepackage{wrapfig}
\usepackage{xspace}
\usepackage{bm}
\usepackage{xcolor}
\usepackage{enumitem}
\usepackage{url}
\usepackage{multirow}
\usepackage{graphicx}
\usepackage{algorithm,algorithmic}
\usepackage{subcaption}
\usepackage{graphicx}
\definecolor{lightgrey}{rgb}{0.43,0.43,0.43}
\definecolor{crimson}{rgb}{0.86,0.08,0.24}
\hypersetup{
  colorlinks,
  citecolor=lightgrey,
  linkcolor=crimson,
  urlcolor=crimson
 }

\makeatletter
\DeclareRobustCommand\onedot{\futurelet\@let@token\@onedot}
\def\@onedot{\ifx\@let@token.\else.\null\fi\xspace}

\def\eg{\emph{e.g}\onedot} 
\def\ie{\emph{i.e}\onedot}

\def\aka{a.k.a\onedot}

\def\wrt{w.r.t\onedot}

\makeatother

\makeatletter
\newcommand*{\centerfloat}{%
  \parindent \z@
  \leftskip \z@ \@plus 1fil \@minus \textwidth
  \rightskip\leftskip
  \parfillskip \z@skip}
\makeatother

\newcommand{\RNum}[1]{\uppercase\expandafter{\romannumeral #1\relax}}

\title{GraphPNAS: Learning Distributions of \\ Good Neural Architectures via Deep \\ Graph Generative Models}

\author{
    Muchen Li$^{1,3}$,\quad Jeffrey Liu$^{2}$,\quad Leonid Sigal$^{1,3,4,5}$,\quad Renjie Liao$^{1}$\\
    \texttt{\{muchenli,lsigal\}@cs.ubc.ca} \quad\texttt{rjliao@ece.ubc.ca}\\
    \texttt{jeffrey.yunfan.liu@uwaterloo.ca}\\
    $^1$University of British Columbia ~~~~~~~ $^2$University of Waterloo\\
    $^3$Vector Institute for AI  ~~~~~~
    $^4$CIFAR AI Chair ~~~~~~ $^5$NSERC CRC Chair
}

\newcommand{\renjie}[1]{{\color{orange}{\bf\sf [Renjie: #1]}}}
\newcommand{\muchen}[1]{{\color{blue}{\bf\sf [Muchen: #1]}}}

\newcommand{\leon}[1]{{\color{red}{\bf\sf [Leon: #1]}}}

\newcommand{\ourmodelshort}{GraphPNAS}

\iclrfinalcopy % Uncomment for camera-ready version, but NOT for submission.
\begin{document}

\maketitle

% \begin{abstract}
% % Neural architecture search has ...
% % We are the first to apply probabilistic graph generator to neural architecture search. 
% % Neural architecture search has been used to 
% Aiming to search for powerful network architectures, neural architecture search(NAS) has been largely focusing on searching for optimal hyper-parameter choice for layers. While previous NAS typically assumes the network architecture to be a chain graph, the searching for neural wiring (the connectivity between layers), are much less explored.
% % the searching for wirings of the neural networks are much less explored.
% Previous NAS practice typically use a RNN based controller and treat this as an sequential decision tasks, which is inefficient in learning and tends to ignore the topological information of neural networks.
% %
% In this paper, we propose to use a graph generative model as a generator to explore neural architecture design space. Unlike commonly used deterministic NAS algorithms, our generator is a pobabilistic model which learns a distribution of promising neural architectures on the target search space. 
% To testify the effectiveness of our graph generator for NAS, we conduct extensive experiments over three datasets with different evaluator \& search spaces, where we demonstrate our proposed graph generator is able to consistently outperform previous methods.  
% \end{abstract}

\vspace{-0.2in}
\begin{abstract}
Neural architectures can be naturally viewed as computational graphs.
Motivated by this perspective, we, in this paper, study neural architecture search (NAS) through the lens of learning random graph models.
In contrast to existing NAS methods which largely focus on searching for a single best architecture, \ie, point estimation, we propose \emph{{\ourmodelshort}}, a deep graph generative model that learns a distribution of well-performing architectures.
% Neural architectures can be naturally viewed as computational graphs.
% Motivating from this perspective, we in this paper study neural architecture search (NAS) from the lens of probabilistic models of graphs.
% In contrast to existing NAS methods which largely focus on searching for a single best architecture, \ie, point estimation, we propose \emph{{\ourmodelshort}}, a deep graph generative model that learns a distribution of well-performing architectures.
Relying on graph neural networks (GNNs), our {\ourmodelshort} can better capture topologies of good neural architectures and relations between operators therein. 
% Relying on graph neural networks (GNNs), \muchenmodi{TODO: justify good property, which take the graph representation of neural architectures as input}, our {\ourmodelshort} better leverages topologies of neural architectures and relations between operators therein. \leon{Can we better characterize in which ways topologies are being leveraged?}
% \muchen{see above}
Moreover, our graph generator leads to a learnable probabilistic search method that is more flexible and efficient than the commonly used RNN generator and random search methods.
% \renjie{You can add one more contribution about search space here!}
% At last, we evaluate architectures under low-data regime so that our generator can be efficiently learned through reinforcement learning.
Finally, we learn our generator via an efficient reinforcement learning formulation for NAS.
% \muchenmodi{At last, we fit our generator to a NAS system where our generator can be efficiently learned through reinforcement learning.}
% \muchenmodi{With the proposed graph generator, we are the first method to search on RANDWIRE: an under-explored challenging search space focusing on the wiring of network layers, where our \ourmodelshort is able to boost the mean accuracy by 0.7\% and reduce the var.}
% \muchenmodi{With {\ourmodelshort}, we are the first method that search for neural wirings on the RANDWIRE search space, a challenging search space with over $10^{149}$ valid architecutres.}
% \muchenmodi{To testify {\ourmodelshort}'s effectiveness over different  NAS setup, we also conduct extensive experiments on ENAS-MACRO and NAS-Bench-101 search space, while we observe consistent performance gain with a\% and b\% respectively .}
To assess the effectiveness of our {\ourmodelshort}, we conduct extensive experiments on three search spaces, including the challenging RandWire on Tiny-ImageNet, ENAS on CIFAR10, and NAS-Bench-101/201.
The complexity of RandWire is significantly larger than other search spaces in the literature.
% which is first explored with our method in the context of NAS.
We show that our proposed graph generator consistently outperforms RNN based one and achieves better or comparable performances than state-of-the-art NAS methods. 
% while up to $14$\% error and $217$\% variance reduction over the three search space and different evaluators.
% 0.7\%$\sim$11.5\% error and 130\%$\sim$188\% variance reduction over the three search space and different evaluators.
% where we demonstrate our proposed graph generator is able to consistently outperform previous methods.  
% \renjie{This last sentence needs to be improved based on your results.}
\end{abstract}

% \muchen{I want to avoid low-data regime here since the current setup is still problematic judging from my data, It will be otherwise hard if we want to update paper with another setup later in rebuttal.}
% \muchen{General logic for abstract: wiring search space under explored, RNN based controller is ineffecient \& do not explicitly capture network topology.}
\vspace{-0.1in}
\section{Introduction}
\vspace{-0.1in}
\label{sec:intro}
% starter point: senquetial network design?
% starter point: neural architecture search?
% \muchen{how to highlight the probabilistic nature of our sampler?}

In recent years, we have witnessed a rapidly growing list of successful neural architectures that underpin deep learning, \eg, VGG, LeNet, ResNets \citep{he2016deep}, Transformers \citep{dosovitskiy2020image}. 
Designing these architectures requires researchers to go through time-consuming trial and errors. 
Neural architecture search (NAS) \citep{zoph2016neural,elsken2018neural} has emerged as an increasingly popular research area which aims to automatically find state-of-the-art neural architectures without human-in-the-loop.

NAS methods typically have two components: a search module and an evaluation module.
The search module is expressed by a machine learning model, such as a deep neural network, designed to operate in a high dimensional search space. 
The search space, of all admissible architectures, is often designed by hand in advance.
The evaluation module takes an architecture as input and outputs the reward, \eg, performance of this architecture trained and then evaluated with a metric.
The learning process of NAS methods typically iterates between the following two steps.
1) The search module produces candidate architectures and sends them to the evaluation module;
2) The evaluation module evaluates these architectures to get the reward and sends the reward back to the search module.
Ideally, based on the feedback from the evaluation module, the search module should learn to produce better and better architectures. 
Unsurprisingly, this learning paradigm of NAS methods fits well to reinforcement learning (RL). 

Most NAS methods \citep{liu2018darts, white2020local,cai2018proxylessnas} only return a single best architecture (\ie, a point estimate) after the learning process.
This point estimate could be very biased as it typically underexplores the search space.
Further, a given search space may contain multiple (equally) good architectures, a feature that a point estimate cannot capture. 
Even worse, since the learning problem of NAS is essentially a discrete optimization where multiple local minima exist, many local search style NAS methods \citep{ottelander2020local} tend to get stuck in local minima.
From the Bayesian perspective, modelling the distribution of architectures is inherently better than point estimation, \eg, leading to the ability to form ensemble methods that work better in practice. 
Moreover, modelling the distribution of architectures naturally caters to probabilistic search methods which are better suited for avoiding local optima, \eg, simulated annealing. 
Finally, modeling the distribution of architectures allows to capture complex structural dependencies between operations that characterize good architectures capable of more efficient learning and generalization.
% \leon{Tried to capture what Ranjie said below}

\begin{comment}
\renjie{
\begin{itemize}
    \item From Bayesian perspective, modelling distribution of architectures is better than point estimation, \eg, model ensemble works better in practice
    \item Modelling distribution of architectures helps construct a learnable probabilistic search method. We know probabilistic search often works better than deterministic local search in discrete optimization, \eg, simulated annealing. A learnable probabilistic search is more flexible since the search strategy is data adaptive.
    \item Modelling distribution of graphs allow easier incorporation of prior on good architectures. 
\end{itemize}
}
\end{comment}

Motivated by the above observations and the fact that neural architectures can be naturally viewed as attributed graphs, we propose a probabilistic graph generator which models the distribution over good architectures using graph neural networks (GNNs).
Our generator excels at generating topologies with complicated structural dependencies between operations. 
From the Bayesian inference perspective, our generator returns a distribution over good architectures, rather than a single point estimate, allowing to capture the multi-modal nature of the posterior distribution of good architectures and to effectively average or ensemble architecture (sample) estimates.
Different from the Bayesian deep learning \citep{neal2012bayesian,blundell2015weight,gal2016dropout} that models distributions of weights/hidden units, we model distributions of neural architectures.
Lastly, our probabilistic generator is less prone to the issue of local minima, since multiple random architectures are generated at each step during learning.
In summary, our key contributions are as below.
\begin{itemize}
    \item We propose a GNN-based graph generator for neural architectures which empowers a learnable probabilistic search method. To the best of our knowledge, we are the first to explore learning deep graph generative models as generators in NAS.
    \item We explore a significantly larger search space (\eg, graphs with 32 operators) than the literature (\eg, garphs with up to 12 operators) and propose to evaluate architectures under low-data regime, which altogether boost effectiveness and efficiency of our NAS system.
    \item Extensive experiments on three different search spaces show that our method consistently outperforms RNN-based generators and is slightly better or comparable to the state-of-the-art NAS methods. Also, it can generalize well across different NAS system setups. 
    % \renjie{We may edit this part once experiments are fixed.}
\end{itemize}

\vspace{-0.1in}
\section{Related Works}
\vspace{-0.1in}
\label{sec:rela}
{\bf Neural Architecture Search.}
The main challenges in NAS are 1) the hardness of discrete optimization, 2) the high cost for evaluating neural networks, and 3) the lack of principles in the search space design. 
First, to tackle the discrete optimization, evolution strategies (ES) \citep{ElskenMH19,RealAHL19}, reinforcement learning (RL) \citep{BakerGNR17,ZhongYWSL18,pham2018efficient,liu2018progressive}, Bayesian optimization \citep{BergstraYC13,white2019bananas} and continuous relaxations \citep{liu2018darts} have been explored in the literature.
We follow the RL path as it is principled, flexible in injecting prior knowledge, achieves the state-of-the-art performances \citep{tan2019efficientnet}, and can be naturally applied to our graph generator.
Second, the evaluation requires training individual neural architectures which is notoriously time consuming\citep{zoph2016neural}.
\citet{pham2018efficient,liu2018darts} propose a weight-sharing supernet to reduce the training time. 
\citet{BakerGRN18} use a machine learning model to predict the performance of fully-trained architectures conditioned on early-stage performances.
\citet{BrockLRW18,zhang2018graph} directly predict weights from the search architectures via hypernetworks. 
% For simplicity and scalability, we propose the strategy of training architectures fully in the low-data regime (fewer samples per class) and then evaluating them.
Since our graph generator do not relies on specific choice of evaluation method, we choose to experiment on both oracle training(training from scratch) and supernet settings for completeness.
Third, the search space of NAS largely determines the optimization landscape and bounds the best-possible performance.
It is obvious that the larger the search space is, the better the best-possible performance and the higher the search cost would likely be. 
Besides this trade-off, few principles are known about designing the search space. 
Previous work \citep{pham2018efficient,liu2018darts,ying2019bench,liwei2020neural} mostly focuses on cell-based search space. 
A cell is defined as a small (\eg, up to 8 operators) computational graph where nodes (\ie, operators like 3$\times$3 convolution) are connected following some topology.
Once the search is done, one often stacks up multiple cells with the same topology but different weights to build the final neural network. 
Other works \citep{tan2019mnasnet,cai2018proxylessnas,tan2019efficientnet} typically fix the topology, \eg, a sequential backbone, and search for layer-wise configurations (\eg, operator types like 3$\times$3 vs. 5$\times$5 convolution and number of filters).
% In our work, we instead consider a much large cell space (32 operators) and search for both operator types and their connectivities. 
In our method, to demonstrate our graph generator's ability in exploring large topology search space, we first explore on a challenging large cell space (32 operators), after which we experiment on ENAS Macro \citep{pham2018efficient} and NAS-Benchmark-101\citep{ying2019bench} for more comparison with previous methods.

{\bf Neural Architecture as Graph for NAS.} 
% Neural architectures can be naturally view as graph and encoded using Graph Neural Networks(GNN) for NAS tasks
{\color{black} Recently, a line of NAS research works propose to view neural architectures as graphs and encode them using graph neural networks (GNNs). In \citep{zhang2020differentiable, luo2018nao}, graph auto-encoders are used to map neural architectures to and back from a continuous space for gradient-based optimization. \citet{shi2020bonas} use bayesian optimization (BO), where GNNs are used to get embedding from neural architectures. Despite the extensive use of GNNs as encoders, few works focus on building graph generative models for NAS.}
Closely related to our work, \citet{xie2019exploring} explore different topologies of the similar cell space using non-learnable random graph models. \citet{you2020graph} subsequently investigate the relationship between topologies and performances.
% Rebuttal Revision
%
{\color{black} Following this, \citet{ru2020nago} propose a hierarchical search space modeled by random graph generators and optimize hyper-parameters using BO.} 
They are different from our work as we learn the graph generator to automatically explore the cell space.
%
% \vspace{-0.1in}

{\bf Deep Graph Generative Models.}
Graph generative models date back to the Erdős–Rényi model \citep{erdos59random}, of which the probability of generating individual edges is the same. 
Other well-known graph generative models include the stochastic block model \citep{holland1983stochastic}, the small-world model \citep{watts1998collective}, and the preferential attachment model \citep{barabasi1999emergence}. 
Recently, deep graph generative models instead parameterize the probability of generating edges and nodes using deep neural networks in, \eg, the auto-regressive fashion \citep{li2018learning,you2018graphrnn,liao2019efficient} or variational autoencoder fashion \citep{kipf2016variational,grover2018graphite,liu2019graph}. 
These models are highly flexible and can model complicated distributions of real-world graphs, \eg, molecules~\citep{jin2018junction}, road networks \citep{chu2019neural}, and program structures \citep{brockschmidt2018generative}.
Our graph generator builds on top of the state-of-the-art deep graph generative model in \citep{liao2019efficient} with several important distinctions.
First, instead of only generating nodes and edges, we also generate node attributes (\eg, operator types in neural architectures).
Second, since good neural architectures are actually latent, our learning objective maximizes the expected reward (\eg, validation accuracies) rather than the simple log likelihood, thus being more challenging.

\vspace{-0.1in}
\section{Methods}
\vspace{-0.1in}
\label{sec:method}
% In this section, we introduce our NAS system.
% % Here we introduce our deep graph generative models that learn a distribution of good architectures through a neural architecture search system.
%
% To shed light on the motivation, we first elucidate the graph representation of neural architectures.
%
% Then we delve into our probabilistic graph generator, evaluator, and the learning method.

% \vspace{-0.05in}
% TODO: Maybe move this to some later part
% Unlike previous works which search and assemble computation cells sequentially to build target network, we focus on a more general search space design where the network(or a computation stage) is encoded by a single DAG. 

\begin{figure}[!t]
  \centering
  \includegraphics[width=0.98\textwidth]{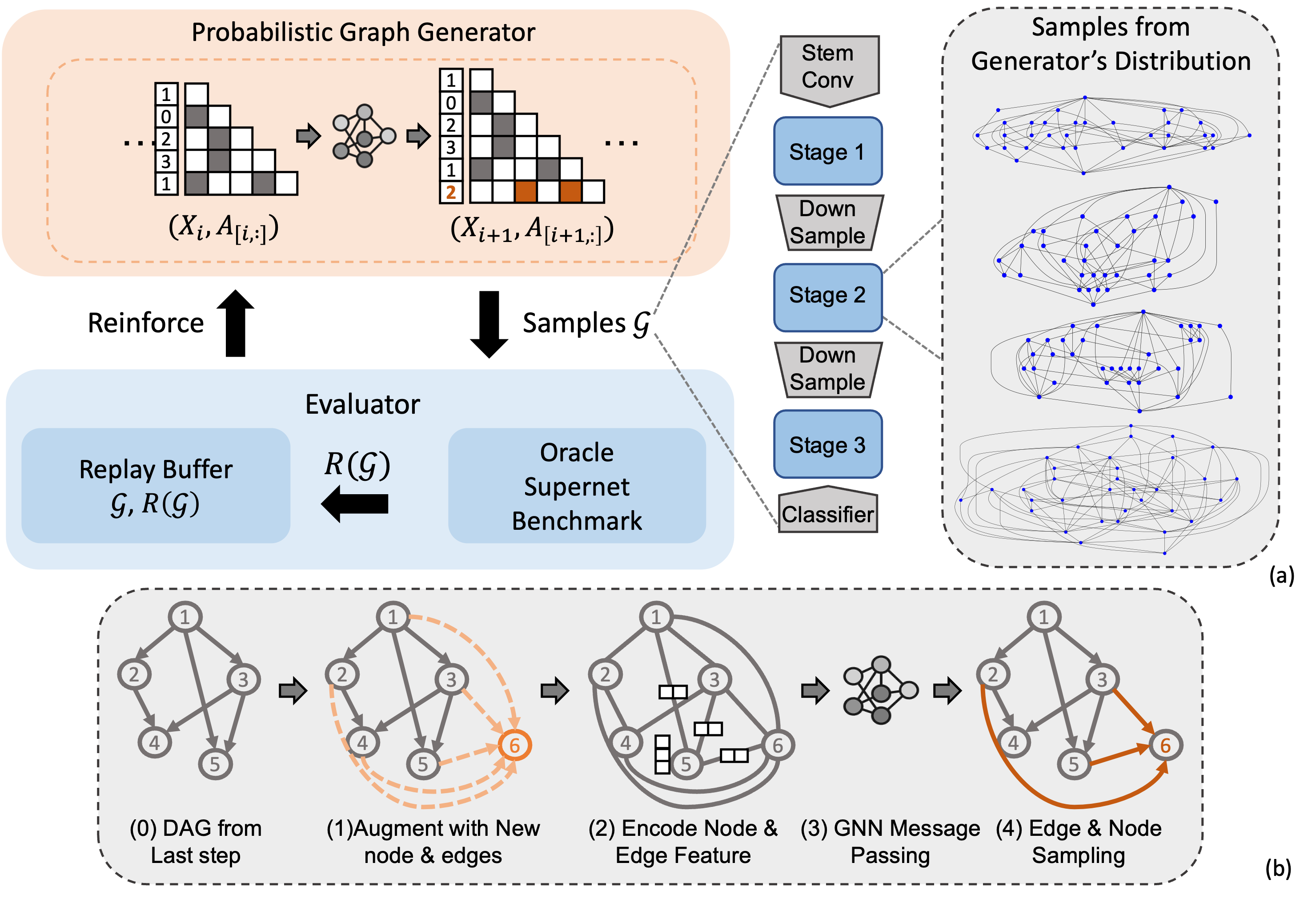}
  \vspace{-0.01in}
  \caption{Figure (a) is the pipeline of our NAS system. The core part is a GNN-based graph generator from which we sample graph representations of neural network $\mathcal{G}$.  The corresponding model for each $\mathcal{G}$ is then sent to evaluator for evaluation. The evaluation result is first stored in a replay buffer and then used for learning the graph generator through Reinforcement Learning.{Figure(b) shows one generation step in the proposed probabilistic graph generator.}
%   The generator samples architecture in an auto-regressive manner: 1: get DAG from last step. 2: Augment it with new node and edges. 3: Encode Graph with node & edge feature, 4: Send graph to GNN for message passing, 5: Sample augmented edge & node to get the new DAG.
  \vspace{-0.1in}
  }
  \label{fig:nas_sys} 
\vspace{-0.05in}
\end{figure}

% \subsection{Neural Architectures As Graphs}
The architecture of any feedforward neural network can be naturally represented as a directed acyclic graph (DAG), \aka, \emph{computational graph}.
There exist two equivalent ways to define the computational graph.
First, we denote operations (\eg, convolutions) as nodes and denote operands (\eg, tensors) as edges which indicate how the computation flows.
Second, we denote operands as nodes and denote operators as edges.
We adopt the first view.
In particular, a neural network $\mathcal{G}$ with $N$ operations is defined as a tuple $(A, X)$ where $A \in \{0,1\}^{N \times N}$ is an $N \times N$ adjacent matrix with $A_{ij}=1$ indicates that the output of the $j$-th operator is used as the input of the $i$-th operator.
For operator with multiple inputs, inputs are combined together  (\eg, using {\tt sum} or {\tt average} operator) before sending into the operator.
$X$ is a $N$-size {\em attribute} vector encoding operation types.
For any operation $i$, its operation type $X_i$ can only choose from a pre-defined list with length $D$, \eg, $1 \times 1$, $3 \times 3$ or $5 \times 5$ convolutions.
% Formally, we have 
% %
% \begin{align}\label{eq:G_def}
%     O_{i} &= \mathbf{L}_{X_i}(\mathbf{F}(\{O_j|A_{ij} = 1\}))
% \end{align}
% % \muchen{clear definition of node and edges, node-> edge ->, operator}
% Here, $D$ is the total number of operations to choose from for each layer, $\mathbf{L}_{X_i}$, and $O_i$ refers to the operation type and the output of the $i$-th layer. $\mathbf{F}$ is the operation we choose to fuse this information ({\em e.g.}, sum, 1d convolution). 
% For example, ResNet \citep{he2016deep} can viewed as a chain graph with $A_{i,i-2:i}=1$ while $N_i$ is $3\times3$ convolution and $\mathbf{F}$ is sum. 
Note that for any valid feedforward architecture, $\mathcal{G}$ can not have loops.
One sufficient condition to satisfy the requirement is to constrain $A$ to be a lower triangular matrix with zero diagonal (\ie, excluding self-loops).
% Note that this construction implicitly assumes that all input and output tensors are of consistent dimensions and can be combined (\eg, using {\tt sum} or {\tt average} operator) to form inputs for nodes that have multiple incoming edges (parents). 
% If the input and the output nodes are given, we treat them as the first and last ones in the row ordering of the adjacency matrix.
% \leon{Do we need the previous sentence?}
% Otherwise, we add an input node representing the input tensor to the whole graph and edges connecting from this node to all nodes without incoming edges (\ie, no parents).
% Similarly, we add an output node and edges connecting from all nodes without out-going edges (\ie, no children) to the output node.
% The final output is the average over all tensors on in-coming edges of the output node.
This formalism creates a search space of $D^N 2^{N(N-1)/2}$ possible architectures, which is huge even for moderately large number of operators $N$ and number of operation types $D$.
% If no constrain is applied to $\mathcal{G}$, we got $D^N2^{N(N-1)/2}$ valid $(A,X)$ pairs, which correspond to the full set of all possible network in this search space. 
The goal of NAS is to find an architecture or a set of architectures within this search space that would perform well. 
For practical consideration, we search for cell graphs (\eg, $N=32$) and then replicate this cell several times to build a deep neural architecture. We also experiment on the ENAS Macro search space where $\mathcal{G}$ defines a entire network.
More details for the corresponding search spaces can be found in Section \ref{sec:exp}.

% \leon{Continue with overview}

% Given no additional constrain for A and X, this setup gives us a search space with the size of $D^N2^{N(N-1)/2}$., which is considerably large when scaling up $D$. 
%, \ie, the space of admissible neural architectures (\aka, the search space in the context of NAS).

\subsection{Neural Architecture Search System}

Before delving into details, we first give an overview of our NAS system, which consists of two parts: a generator and an evaluator.
The system diagram is shown in Fig. \ref{fig:nas_sys}. 
At each step, the probabilistic graph generator samples a set of cell graphs, which are further translated to neural architectures by replicating the cell graph multiple times and stacking them up. 
Then the evaluator evaluates these architectures, obtains rewards, and sends architecture-reward pairs to the replay buffer. The replay buffer is then used to improve the generator, effectively forming a reinforcement learning loop. 

\vspace{-0.5em}
\subsubsection{Probabilistic Generators for Neural Architectures}\label{sec:generator}
Now we introduce our probabilistic graph generator which is based on a state-of-the-art deep auto-regressive graph generative model in \citep{liao2019efficient}.

{\bf Auto-Regressive Generation.\quad}
Specifically, we decompose the distribution of a cell graph along with attributes (operation types) in an auto-regressive fashion,
\begin{align}\label{eq:auto_regressive_model}
    \mathbb{P} (A, X) = \prod_{i=1}^{N} \mathbb{P} \left(A_{i,:} \middle| A_{i-1,:}, X_{i-1}, \cdots, A_{1,:}, X_{1} \right) \mathbb{P} \left(X_{i} \middle| A_{i-1,:}, X_{i-1}, \cdots, A_{1,:}, X_{1} \right),
\end{align}
where $A_{i,:}$ and $X_{i}$ denote the $i$-th row of the adjacency matrix $A$ and the $i$-th operation type respectively.
To ensure the generated graphs are DAGs, we constrain $A$ to be lower triangular by adding a binary mask, \ie, the $i$-th node can only be reached from the first $i-1$ nodes.
We omit the masks in the equations for better readability.
We further model the conditional distributions as follows,
\begin{align}\label{eq:conditional_distribution}
    \mathbb{P} \left(A_{i,:} \middle| A_{i-1,:}, X_{i-1}, \cdots, A_{1,:}, X_{1} \right) & = \sum_{k=1}^{K} \alpha_{k} \prod_{1 \le j < i} \theta_{k,i,j} \\
    \mathbb{P} \left(X_{i} \middle| A_{i-1,:}, X_{i-1}, \cdots, A_{1,:}, X_{1} \right) & = \text{Categorical}\left( \beta_{1}, \cdots, \beta_{D} \right) \\ \label{eq:conditional_distribution4}
    \alpha_{1}, \dots, \alpha_{K} & = \text{Softmax} \left( \sum\nolimits_{1 \le j < i} \text{MLP}_{\alpha}(h^S_i - h^S_j) \right) \\
    \label{eq:conditional_distribution5}
    \beta_{1}, \dots, \beta_{D} & = \text{Softmax} \left( \text{MLP}_{\beta}(h^S_i) \right) \\
    \label{eq:conditional_distribution6}
    \theta_{1,i,j}, \dots, \theta_{K,i,j} & = \text{Sigmoid} \left( \text{MLP}_{\theta}(h^S_i - h^S_j) \right),
\end{align}
where the distributions of the operation type and edges are categorical and $K$-mixture of Bernoulli respectively.
$D$ is again the number of operation types.
$\text{MLP}_{\alpha}$, $\text{MLP}_{\beta}$, and $\text{MLP}_{\theta}$ are different instances of two-layer MLPs with ReLU activations.
Here $h^S_i$ is the representation of $i$-th node returned by a GNN which has been executed $S$ steps of message passing at each generation step.
This auto-regressive construction breaks down the nice property of permutation invariance for graph generation.
However, we do not find it as an issue in practice, partly due to the fact that the graph isomorphism becomes less likely to happen while considering both topology and operation types. 

{\bf Message Passing GNNs.\quad} 
Each generation step $n \le N$ in auto-regressive generation above relies on representations of nodes up to and including $n$ itself (see Eq.~(\ref{eq:conditional_distribution4})--(\ref{eq:conditional_distribution6})). To obtain these node representations $\{h^S_i\}$, we exploit message passing GNNs \citep{gilmer2017neural} with an attention mechanism similar to \citep{liao2019efficient}. 
In particular, the $s$-th message passing step involves executing the following equations successively, \\
\begin{minipage}{.5\linewidth}
    \centering
    \begin{align}
        m^s_{ij} & = f ( [h^s_{i} - h^s_{j}, \bm{1}_{ij}]) \label{eq:msg} \\
        \tilde{h}^s_{i} & = [h^s_{i}, u_i]  \label{eq:pad_hidden}
    \end{align}
\end{minipage}%
\hfill
\begin{minipage}{.5\linewidth}
    \centering
    \begin{align}
        a^s_{ij} & = \text{Sigmoid} ( g ( \tilde{h}^s_{i} - \tilde{h}^s_{j})) \label{eq:attention} \\
        h^{s+1}_{i} & = \text{GRU} ( h^s_{i}, \sum\nolimits_{j \in \mathcal{N}(i)} a^s_{ij} m^s_{ij} ). \label{eq:state_update}
    \end{align}
\end{minipage}
where $\mathcal{N}(i)$ is the set of node $i$ along with its neighboring nodes. 
$m^s_{ij}$ is the message sent from node $i$ to $j$ at the $s$-th message passing step.
The connectivity for the propagation in GNN is given by $A_{1:i-1,1:i-1}$ with the last node (for which $A_{i,:}$ has not been generated yet) being fully connected. 
Note that message passing step is different from the generation step and we run multiple message passing steps per generation step in order to capture the structural dependency among nodes and edges.
The $f$ and $g$ are two-layer MLPs.
Since graphs are DAGs in our case rather than undirected ones as in \citep{liao2019efficient}, we add $\bm{1}_{ij}$ in Eq. (\ref{eq:msg}), a one-hot vector for indicating the direction of the edge.
We initialize the node representations $h^0_i$ (for $i < n$) as the corresponding one-hot encoded operation type vectors; % padded with zeros; 
$h^0_n$ is initialized to a special one-hot vector.
Here $u_i$ is an additional feature vector that helps distinguish $i$-th node from others. 
We found using one-hot-encoded incoming neighbors of $i$-th node and a positional encoding of the node index $i$ work well in practice. We encourage readers to reference Fig. \ref{fig:graph_generation} for a detailed visualization of graph generation process.
% Note that additional features are not permutation equivariant, but we find that adding these features empirically helps the learning of our graph generator.

{\bf Sampling. \quad}To sample from our generator, we first draw architectures following the standard ancestral sampling where each step involves drawing random samples from a categorical distribution and a mixture of Bernoulli distribution.
At each step, this sampling process adds a new operator with a certain operation type and wire it to previously sampled operators.
% We use the rejection sampling to reject architecture that is invalid in the corresponding search space.
% \renjie{You need to explicitly explain the criterion of rejection here!}
\vspace{-0.5em}
\subsubsection{Evaluator}
Our design of generator and NAS pipeline do not rely on a specific choice of evaluator.
Motivated by \citep{mnih2013playing}, we use a replay buffer for storing the evaluated architectures.
In our paper, based on specific datasets, we explore three types of evaluators, namely, oracle evaluator, supernet evaluator and benchmark evaluator, which are briefly introduced as follows.

{\bf Oracle evaluator.} Given a sample from the generator, an oracle evaluator trains the corresponding network from scratch and tests it to get the validation performances. 
% To reduce the computation overhead, one can either choose to use early-stopping (training with fewer epochs) as in \citep{tan2019mnasnet} or employ a fast proxy (\eg, downsample dataset/reduce model size) \citep{dong2020bench}. 
To reduce computation overhead, a common approach is to use early stopping (training with fewer epochs) as in \citep{tan2019mnasnet,tan2019efficientnet}.
In our experiment, we instead use a low-data evaluator similar to few-shot learning where we keep the same number of classes but use fewer samples per class to train. %  the model.
% is used while keeping the number of training epochs. 
% We show in the experiments section that early stopping can lead to local optima given the unique challenge brought by the search space.

{\bf SuperNet evaluator.} Aiming at further reducing the amount of compute, this evaluator uses a weight-sharing strategy where each graph is a sub-graph of the supernet. 
We followed the single-path supernet setup used in \citep{pham2018efficient} to compare with previous methods.
% in which a node of the same operation type with the same id shares weight. During the supernet training / testing, only a single path is sampled.
% We follow this setup to fairly compare with ENAS.

{\bf Benchmark evaluator.} NAS benchmarks, \eg, \citep{ying2019bench}, provide accurate evaluation for architectures within the search space, which can be seen as oracle evaluators with full training budgets on target datasets.
% The Benchmark evaluator can be seen as a an oracle evaluator without bias

%  \muchen{one more sentence here}
% \renjie{Add a few sentences to briefly explain three types of evaluators and then spend some space to explain our low-data regime based evaluator.}

\vspace{-0.5em}
\subsection{Learning Method}
Since we are dealing with discrete latent variables, \ie, good architectures in our case, we train our NAS system using REINFORCE \citep{williams1992simple} algorithm with the control variate (\aka baseline) to reduce the variance.
In particular, the gradient of the loss or negative expected reward $\mathcal{L}$ \wrt the generator parameters $\phi$ is,
\begin{align}\label{eq:gradient_reinforce_baseline}
	\nabla \mathcal{L}(\phi) = \mathbb{E}_{\mathbb{P}(\mathcal{G})} \left[- \frac{\partial \log \mathbb{P}(\mathcal{G})}{\partial \phi} \bar{R}(\mathcal{G}) \right],
\end{align}
where the reward $\bar{R}$ is standardized as $\bar{R}(\mathcal{G}) = (R(\mathcal{G}) - C)/\sigma$.
Here the baseline $C$ is the average reward of architectures in the replay buffer and
$\sigma$ is standard deviation of rewards in the replay buffer.
The expectation in Eq. (\ref{eq:gradient_reinforce_baseline}) is approximated by the Monte Carlo estimation.
However, the score function (\ie, the gradient of log likelihood \wrt parameters) in the above equation may numerically differ a lot for different architectures.
For example, if a negative sample, \ie, an architecture with a reward lower than the baseline, has a low probability $\mathbb{P}(\mathcal{G})$, it would highly likely to have an extremely large absolute score function value, thus leading to a negative reward with an extremely large magnitude.
Therefore, in order to balance positive and negative rewards, we propose to use the reweighted log likelihood as follows,
\begin{align}\label{eq:reweight_nll}
    \log \mathbb{P} (\mathcal{G}) = \beta \mathbf{1}_{\bar{R}(\mathcal{G}) \leq 0} \log(1 - \mathbb{P}(\mathcal{G})) + \mathbf{1}_{\bar{R}(\mathcal{G}) > 0} \log(\mathbb{P}(\mathcal{G})) 
\end{align}
where $\beta$ is a hyperparameter that controls the weighting between negative and positive rewards.
$\mathbb{P}(\mathcal{G})$ is the original probability given by our generator.

{\bf Exploration vs. Exploitation}
Similar to many RL approaches, our NAS system faces the exploration vs. exploitation dilemma.
We found that our NAS system may quickly collapse (\ie, overly exploit) to a few good architectures due to the powerful graph generative model, thus losing the diversity and reducing to point estimate.
Inspired by the epsilon greedy algorithm \citep{sutton2018reinforcement} used in multi-armed bandit problems, we design a random explorer to encourage more exploration in the early stage.
Specifically, at each search step, our generator samples from either itself or a prior graph distribution like the Watts–Strogatz model with a probability $\epsilon$.
As the search goes on, $\epsilon$ is gradually annealed to 0 so that the generator gradually has more exploitation over exploration.
{\color{black} Whats more, we design our replay buffer to keep a small portion of candidates. As training goes on, bad samples will be gradually be replaced by good samples for training our generators, which encourage the model to exploit more.}

\vspace{-0.1in}
\section{Experiments}
\vspace{-0.1in}
\label{sec:exp}
% \subsection{Experiments Details}
\begin{comment}
Cell-based search spaces \citep{liu2018darts,ying2019bench} have been extensively researched given it's robust performance lower bound and good transferability when scaling up. There has been little exploration for searching on macro search space where connectivity between different layers is considered.
\end{comment}
%
In this section, we extensively investigate our NAS system on three different search spaces to verify its effectiveness.
\begin{comment}
In NAS literatures\citep{zoph2016neural,li2020random}, RNN based controller and random search methods has been extensively used as the state-the-art generator or strong baseline. In our experiments, we comprehensively compare our graph generator with RNN and random search on three search spaces.
We show that our graph generator consistently outperforms RNN based generator and random search methods on all three search spaces.
\end{comment}
%
First, we adopt the challenging RandWire search space \citep{xie2019exploring} which is significantly larger than common ones.
To the best of our knowledge, we are the first to explore learning NAS systems in this space.
Then we search on the ENAS Macro \citep{pham2018efficient} and NAS-Bench-101\citep{ying2019bench} search spaces to further compare with previous literature.
% \muchen{I want to express the first search space is under-explored and second and third is better studied so we can stably compare with previous methods}
% \subsection{Generator details}
For all experiments, we set the number of mixture Bernoulli $K$ to be 10, the number of message passing steps $S$ to 7, hidden sizes of node representation $h_i^s$ and message $m^s_{ij}$ to 128. 
For RNN-based baselines, we follow the design in \citep{zoph2018learning} if not other specified.

\vspace{-0.5em}
\subsection{RandWire Search Space on Tiny-ImageNet}
\vspace{-0.5em}
\label{sec:randwire_result}
{\bf RandWire Search Space.} Originally proposed in \citep{xie2019exploring}, a randomly wired neural network is a ResNet-like four-stage network with the cell graph $\mathcal{G}$ defines the connectivity of $N$ convolution layers within each stage. 
At the end of each stage, the resolution is downsampled by 3$\times$3 convolution with stride 2 whereas the number of channels is doubled. 
While following the RandWire small regime in \citep{xie2019exploring}, we share the cell graph $G$ among last three stages for simplification.
To keep the number of parameters roughly the same, we fix the node type to be separable 3$\times$3 convolution.
The number of nodes $N$ within the cell graph $\mathcal{G}$ is set to 32 excluding the input and output nodes.
% A $\mathcal{G}$ is considered as valid if all the intermediate nodes are connected to it. 
% \renjie{I am confused by the previous sentence.}
This yields a search space of $2.1\times 10^{149}$ valid adjacency matrices, which is extremely large and renders the neural architecture search challenging. 
More details of the RandWire search space can be found in the Appendix \ref{app:randwire}.

\begin{table}[htbp]
\vspace{-0.1in}
\small
\begin{center}
	\begin{tabular}{l|c|c|c|c|c}
		\hline
		\multirow{2}{*}{Methods} & Cost & \multicolumn{2}{c|}{Low Data (Search)} & \multicolumn{2}{c}{Full Data (Final)}\\ 
		\cline{3-6}
		& (GPU Days) & Val Avg Acc & Std & Val Avg Acc & Std \\
		\hline
        \hline 
		% \textit{Random Search Methods} & & & & & \\
        ER-TopK     & 15.2 & 23.12 & 0.34 & 61.76 & 0.04 \\
        WS-TopK     & 15.6 & 22.39 & 0.91 & 62.24 & 0.34 \\
        ER-BEST     & 15.2 & 20.07 & 1.62 & 62.10 & 0.25 \\
        WS-BEST     & 15.6 & 18.68 & 1.41 & 62.16 & 0.92 \\        
		\hline 
		\hline
		% \textit{Learning-to-Search Methods} & & & & &\\
% 			\multirow{2}{*}{Watts-Strogat} & 10.4 & & & \\
% 			 & 20.9 & & & \\
% 			\multirow{2}{*}{Erdos-Renyi} & 10.4 & & & \\
% 			& 20.9 & & & \\
        % WS, $k,p=4,0.8$& 16.4 & 18.68 & 1.41 & 62.35 & 0.48\\
		RNN \citep{zoph2018learning} & 17.2 & 18.46 & \textbf{0.99} & 61.73 & 0.77\\
		Ours & 16.7 & \textbf{20.32} & 1.12 & \textbf{62.57} & \textbf{0.40}\\
		\hline
	\end{tabular}
\end{center}
\caption{Comparisons on Tiny-ImageNet. The top and bottom blocks include random search and learning-to-search methods respectively. ER-TopK and WS-TopK refers to top ($K$=4) architectures found by all WS and ER models during search. ER-BEST and WS-BEST refer to the best ER and WS models found during search, \ie, WS($k$=4,$p$=0.75) and ER($p$=0.1). Here Avg and Std of accuracies are computed based on 4 architectures sampled from generators.\vspace{-0.25in}}
\vspace{-0.15in}
\label{tab:prob_randwire}
\end{table}

{\bf Tiny-ImageNet w. Oracle Evaluator.} To enable search on the RandWire space, we exploit the oracle evaluator on the Tiny-ImageNet dataset \citep{chrabaszcz2017downsampled}. 
To save computation, we employ a low-data oracle evaluator where we sample $1/10$ of Tiny-ImageNet training set for training and use the rest for validation at each search step. 
Similar to the few-shot learning, we keep the number of classes unchanged but reduce the number of samples per class.
After the search, we retrain our found architectures on the full training set and evaluate it on the original validation set. 
Specifically, for each model, the oracle evaluator trains for 300 epochs and uses the average validation accuracy of the last 3 epochs as the reward. 
Our total search budget is around 16 GPU days, which approximately amounts to 320 model evaluations, \eg, 40 search steps and 8 samples evaluated per step. 
For random search baselines, we choose Erdős–Rényi (ER) and Watts–Strogatz (WS) models. 
Specifically, we first randomly draw hyperparameters from certain ranges, \ie, $0.1\leq p\leq 0.5$ for ER and $(2,0.2)\leq(k,p)\leq (6,0.8)$ for WS, and then sample $\mathcal{G}$ from individual models. 
We set the reweight coefficient $\beta$ to 0.05. 
For the random explorer, we choose WS model with the same hyperparameter range as a prior distribution and set $\epsilon=0.6$ in the beginning and decay it by a factor of 0.2 every 10 search steps. 
We also find that gradually shrinking replay buffer size to keep $30\%$ to $10\%$ of top-performing architectures helps stabilize the training of the generator.
At the search time, we reject samples that already appear in the replay buffer to avoid duplications.
We apply the same setting to the RNN generator for a fair comparison.

\textbf{Results.}
As shown in Table \ref{tab:prob_randwire}, we compare our NAS system with other random search methods and learning-to-search methods. 
We can see that our method outperforms the RNN-based generator and other random search methods in terms of average validation accuracy on the full dataset.
Our generator also has a lower variance compared to the RNN-based one.
Moreover, we observed that RNN based generator sometimes degenerates so that it frequently sample densely-connected graphs. 
This is probably due to the fact that RNN based generator does not effectively utilize the topology information.
%given the long term dependency is hard to learned. 
We can see that a high search reward (\ie, a low-data validation accuracy) do not necessarily lead to better performances in full data training, which indicates a bias of the oracle evaluator within the low-data regime. 
% Note that for random search methods, top architectures with the best validation reward at search time are chosen for evaluations. 
Random search methods are prone to be biased as they select architectures solely based on the search reward. 
Nevertheless, our generator is less affected by the bias and able to learn a distribution of good architectures that perform well on full data training. 

\begin{wraptable}[15]{r}{0.58\textwidth}
\vspace{-0.1in}
% \begin{table}[htbp]
\caption{Comparisons of best searched architectures (averaged over 3 runs per architecture) on Tiny ImageNet. }
\small
\centering
\resizebox{\linewidth}{!}{
    \begin{tabular}{l|c|c|c}
		\hline
		\multirow{1}{*}{Model} & Param (M) & \multicolumn{2}{c}{Top1 / Top5 Acc}\\ 
% 			& (Million) & (Evaluation times) & \% \\
		\hline 
		Resnet18 & 11.68 & $59.71_{\pm0.09}$& $80.32_{\pm0.10}$ \\
		Resnet50 & 25.56 & $63.42_{\pm0.30}$& $82.61_{\pm0.15}$ \\
		Resnext50 & 27.56 & $63.62_{\pm0.07}$& $82.73_{\pm 0.08}$\\
% 			Residual Chain & & & & \\
		\hline
		FC & 3.49 & $60.82_{\pm0.24}$ & $82.29_{\pm 0.09}$\\
		ER-Top1 & 3.23 & $61.82_{\pm0.09}$ & $82.30_{\pm 0.18}$\\
		RS-Top1 & 3.22 & $62.55_{\pm0.15}$ & $82.64_{\pm 0.21}$\\
		RNN & 3.32 &$62.29_{\pm 0.39}$ & $82.16_{\pm0.24}$ \\
		Ours & 3.27 & \textbf{63.23$_{\pm0.18}$} & \textbf{83.06$_{\pm 0.05}$}\\
		\hline
		WS-Top1 Large & 19.38 & $63.84_{\pm0.13}$&  $82.61_{\pm0.16}$\\
		RNN Large & 19.78 & $63.69_{\pm0.28}$& $82.74_{\pm 0.21}$\\
		Ours Large & 19.18 & \textbf{64.45$_{\pm0.26}$} & \textbf{83.23$_{\pm0.26}$} \\
		\hline
	\end{tabular}}
% \end{table}
\label{tab:best_randwire}
\vspace{-0.1in}
\end{wraptable}

We also show results of the best architectures found within 4 samples in Table \ref{tab:best_randwire}.
Here, ER-top-1 and WS-top-1 refers to the best model found from corresponding random search. FC refers to the fully-connected graph, which takes three times longer to train compared to our model.
It is clear that the best model found by our method outperforms those discovered by other methods by a considerable margin. 
Moreover, we scale up the best models (denoted as large) by adding more channels and one more computation stage (more details are in Appendix \ref{app:randwire}). 
We can see that our searched architectures perform favorably against manually designed architectures like ResNet \citep{he2016deep} and ResNeXt \citep{xie2017aggregated}.

%
% All the models here use a C=64, the corresponding model sizes ranges from 3.2 to 3.6M. 
%
% For the choice of oracle evaluator, we also tried to follow \cite{tan2019mnasnet}, where early stopping is used to save computation. We observe that doing this introduces larger bias while all the methods are encouraged to search shallow network given they are faster in converging.
% \leon{last bit is confusing}

\vspace{-0.5em}
\subsection{ENAS Macro Search Space on CIFAR10}
\vspace{-0.5em}
\label{sec:enas_result}
{\bf ENAS Macro Search Space}, originally proposed by \citet{pham2018efficient}, is a search space which focuses on the entire network. 
$\mathcal{G}$ here defines the entire network with $N=12$ nodes. 
The operation type ($D$=6)\footnote{$1 \times 1$, $5 \times 5$ convolution, $1 \times 1$, $5 \times 5$ separable convolution, max pooling, avg pooling} per node is also searchable. 
$\mathcal{G}$ is guaranteed to contain a length-11 path, \ie, $\forall i>1$, $A_{i,i-1}=1$. 
The goal is to search off-diagonal entries, \ie, skip connections. 
This gives a search space of $1.6\times10^{29}$ valid networks in total. 

{\bf SuperNet Evaluator.} For ENAS Macro search space, we experiment on CIFAR10 \citep{cifar} dataset. 
For our generator, we use the ER model with $p=0.4$ as our explorer, where $\epsilon$ decays from 1 to 0 in the first 100 search steps. 
For RNN based generator, we follow the setup in \citep{pham2018efficient}.
We also adopt the weight-sharing mechanism in \citep{pham2018efficient} to obtain a SuperNet evaluator that efficiently evaluates a model's performance. 
We use a budget of 300 search steps with around 100 architectures evaluated per step for all methods. 
After the search, we use a short-training of 100 epochs to evaluate the performances of 8 sampled architectures, after which top-4 performing ones are chosen for a 600-epoch full training. 
The best validation error rate among these 4 architectures is reported.
For simplicity and a fair comparison, we do not use additional tricks (\eg, adding entropy regularizer) in \citep{pham2018efficient}.
More details are provided in Appendix \ref{app:enas}.

\begin{table}[t]
	\caption{Comparisons on CIFAR10 dataset. The bottom and top blocks include NAS methods with ENAS Macro and other search spaces respectively. *: our re-implementation. -: inapplicable.\vspace{-0.1in}
% 	Here Avg and variance are calculated from top 4 performing archs out of 8 samples from the generator.
	}
	\vspace{-0.05cm}
	\small
	\begin{center}
		\begin{tabular}{l|c|c|c|cc}
			\hline
			\multirow{2}{*}{Methods} & Search Cost & Params & Best & \multicolumn{2}{c}{Top Samples}\\ 
			 & (days) & (M) & Error Rate & Avg & Std\\
            \hline
            \hline
            % NAS \citep{zoph2016neural} & 22400 & 7.1 & 95.53 & - & -\\
            Net Transform \citep{cai2018path} & 10 & 19.7 & 5.7 & - & -\\
            NAS \citep{zoph2016neural} & 22400 & 7.1 & 4.47 & - & -\\
            % PNAS \citep{liu2018progressive} & 225 & 3.2 & 96.59 & -& - \\
            PNAS \citep{liu2018progressive} & 225 & 3.2 & 3.41 & -& - \\
            Lemonade \citep{elsken2018efficient} & 56 & 3.4 & 3.6 & - & -\\
		    EPNAS-Macro \citep{perez2018efficient} & 1.2 & 38.8 & 4.01 & - & -\\
			\hline 
            \hline
% 			RNN* \multirow{2}{*}{\cite{pham2018efficient}} & 0.9 & 19.64 & 95.82 & 95.53 & 0.282\\			
			RNN* \citep{pham2018efficient} & 0.9 & 19.64 & 4.18 & 4.47 & 0.282\\
% 			RNN* Large & 0.9 & 36.92 & 96.00 & 95.84 & 0.089\\
			RNN* Large & 0.9 & 36.92 & 4.00 & 4.16 & 0.089\\
% 			Ours & 0.5 & 20.47 & 96.27 & 96.07 & 0.098\\
			Ours & 0.5 & 20.47 & 3.73 & 3.93 & 0.098\\
% 			Ours Large & 0.5 & 37.71 & \textbf{96.45} & \textbf{96.38} & \textbf{0.050}\\
			Ours Large & 0.5 & 37.71 & \textbf{3.55} & \textbf{3.62} & \textbf{0.050}\\
			\hline
		\end{tabular}
	\end{center}
	\label{tab:enas_results}
    \vspace{-0.1in}
\end{table}
In Table \ref{tab:enas_results}, we compare the error rates and variances for different NAS methods.
Note that this variance reflects the uncertainty of the distribution of architectures as it is computed based on sampled architectures.
% We can see that our graph generator reaches an 14\% average error rate reduction and 217\% performance variance reduction. 
It is clear that our {\ourmodelshort} achieves both lower error rates and lower variances compared to RNN based generator and is on par with the state-of-the-art NAS methods on other search spaces.
We also see that the best architecture performance of our generator outperforms RNN based generator by a significant margin. 
This verifies that our {\ourmodelshort} is able to learn a distribution of well-performing neural architectures.
% \muchen{Do we want this sentence: }
Given that we only sample 8 architectures, the performances could be further improved with more computational budgets. 
% one can increase the number of samples for low-budget pre-evaluation to get potential better architectures.

% (\eg, enforce sparsity by adding a KL divergence term in reward) to regularize our controller during training. 
% Yet we still observe stable training and better results.

% \paragraph{Analysis on} 
% \leon{Something still missing here?}

\vspace{-0.5em}
\subsection{NAS Benchmarks}
\vspace{-0.5em}
\label{sec:nasbench_result}
{\bf NAS-Bench-101} \citep{ying2019bench} is a tabulate benchmark containing $423$K cell graphs, each of which is a DAG with up to $7$ nodes and $9$ edges including input and output nodes. 
% \muchen{Hi jeffrey can you check following sentence: } 
%
\begin{figure*}[t]
    \RawFloats
    \centering
    \begin{minipage}{.45\linewidth}
        % \vspace{-0.1in}
        \centering
        \includegraphics[width=\linewidth]{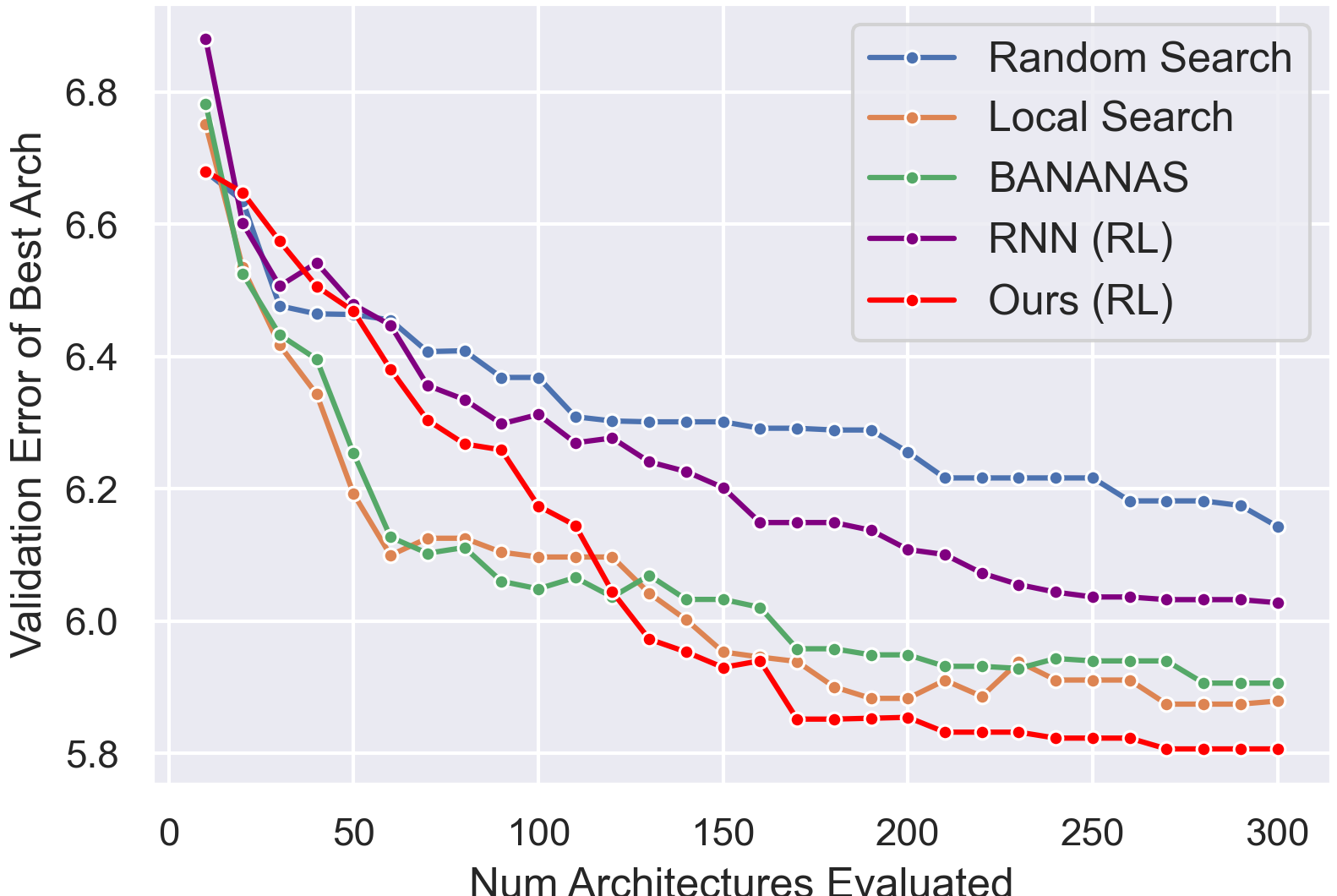}
        \vspace{-0.2in}
        \caption{\color{black} Performances (average over 10 runs) of best architectures vs. the number of architecture evaluations (search step).}
        \label{fig:nasbench}
    \end{minipage}\quad
    \begin{minipage}{.5\linewidth}
        \vspace{-0.2in}
        \begin{table}[H]
            \centering
            \resizebox{0.95\linewidth}{!}{%
                \begin{tabular}{ l|c|c } 
                    \hline
                    Method & Avg Error & \#Queries \\ 
                    \hline
                    \hline
                    % RNN$^{\dagger}$ & 6.436 & 150\\
                    GCN Pred$^{\dagger}$ & 6.331 & 150\\
                    Evolution$^{\dagger}$ & 6.109 & 150\\
                    Ours & 5.930 $_{\pm\text{0.143}}$& 150 \\
                    \hline
                    \hline
                    Random Search & 6.413 $_{\pm\text{0.422}}$& 300\\ 
                    Local Search & 5.879$_{\pm\text{0.371}}$ & 300\\ 
                    BANANAS & 5.906 $_{\pm\text{0.296}}$& 300 \\
                    RNN (RL)& 6.028$_{\pm\text{0.228}}$ & 300 \\
                    Ours (RL)& \textbf{5.807}$_{\pm\textbf{0.072}}$ & 300 \\
                    \hline
                \end{tabular}
            }
            \vspace{0.1in}
            \caption{Best model performances on NAS-Bench-101. $^\dagger$ indicates numbers are taken from \citep{white2019bananas} Table 2.}
            \label{tab:nasbench}
        \end{table}
    \end{minipage}
    \vspace{-0.1in}
    % \caption{some wonderful words.}
\end{figure*}
We compare the performances of our {\ourmodelshort} to open-source implementations of random search methods, local search methods, and BANANAS \citep{white2019bananas}. 
The latter two are the best algorithms found by \citet{white2020local} on NAS-Bench-101. 
For GCN prediction and evolution methods, we use the score reported in \citep{white2020local}. 
% \jeffrey{This is correct. However I only retrain the generator once every 10 samples.}
% \jeffrey{we keep top 30 archs (not percent)}
We give each NAS method the same budget of 300 queries and plot the curve of lowest test error as a function of the number of evaluated architectures. 
% \jeffrey{Perhaps it is more precise to say ``test error'' here} 
As shown in Fig. \ref{fig:nasbench}, our {\ourmodelshort} is able to quickly find well-performing architectures.
We also report the avg error rate over 10 runs in Table \ref{tab:nasbench}. 
Our {\ourmodelshort} again outperforms RNN based generator by a significant margin and beats strong baselines like local search methods and BANANAS. 
Notably, our {\ourmodelshort} has a much lower variance than other methods, thus being more stable across multiple runs.

{\bf NAS-Bench-201} { \citep{dong2020bench} is defined on a smaller search space where up to $4$ nodes and $6$ edges are allowed. Experimental results can be found in Appendix \ref{sec:nasbench_201}}.
% \begin{figure}
% \begin{floatrow}
% \ffigbox{%
%   \includegraphics[width=0.6\textwidth]{imgs/nb101_compare.png}
% }{%
%   \caption{A figure}%
% }
% \capbtabbox{%
%     \begin{tabular}{ |c|c|c| } 
%     \hline
%     Method & Avg Error &  Variance \\ 
%     \hline
%     Random Search & 6.24 & 0.20 \\ 
%     Local Search & 5.93 & 0.15 \\ 
%     BANANAS & 5.90 & 0.14 \\
%     \hline
%     Ours & \textbf{5.79} & \textbf{0.05} \\
%     \hline
%     \end{tabular}
% }{%
%   \caption{A table}%
% }
% \end{floatrow}
% \end{figure}

\vspace{-0.1in}
\section{Discussion \& Conclusion}
\vspace{-0.1in}
\label{sec:discuss}
{\bf Qualitative Comparisons between RNN and {\ourmodelshort}.}
\begin{figure}[t]
    \vspace{-0.1in}
    \centering
    \includegraphics[width=0.8\textwidth]{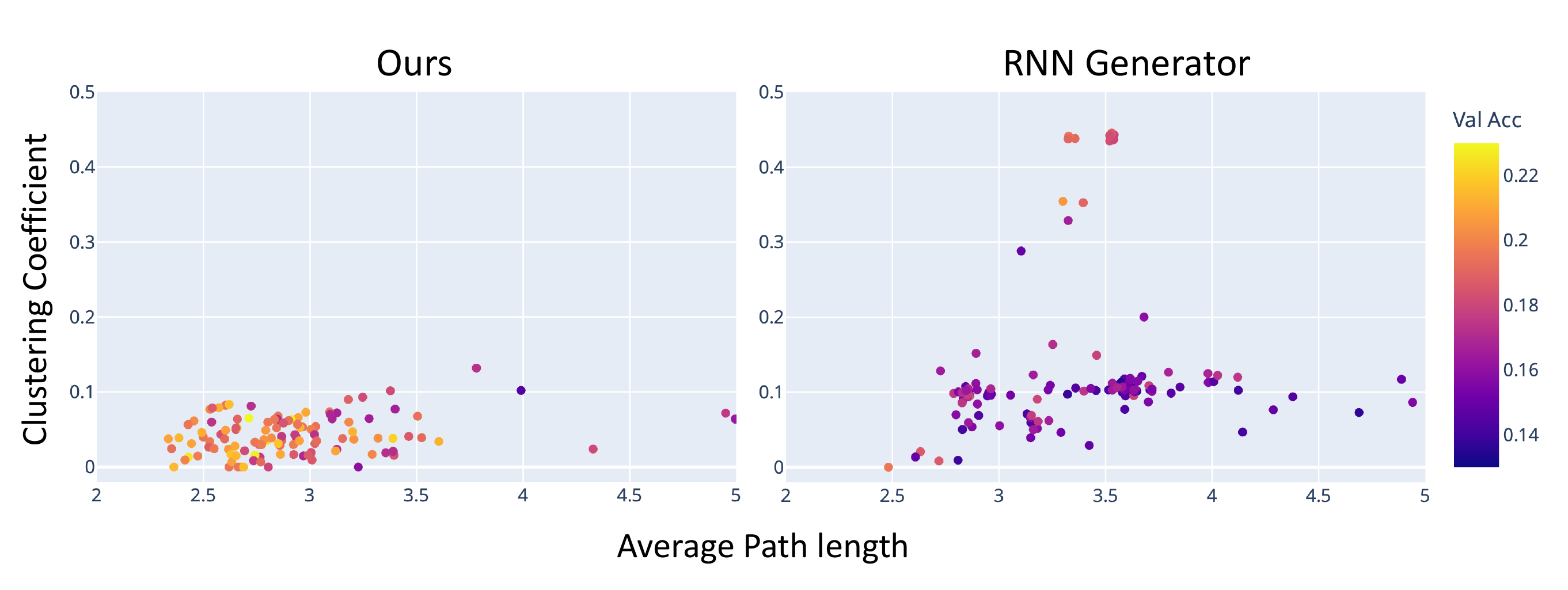}
    \vspace{-0.1in}
    \caption{Visualization of architecutre explore sapce of {\ourmodelshort} vs RNN. Each point in the figure denotes a model evaluation. Colors of each node denotes its validation accuracy returned by the low-data Oracle evaluator. \vspace{-0.25in}}
    \label{fig:compare_explore}
    \vspace{-0.1in}
\end{figure}
In \citep{you2020graph}, the clustering coefficient and the average path length have been used to investigate distributions of graphs. 
Here we adopt the same metrics to visualize architectures (graphs) sampled by RNN based and our generators in RandWire experiments. 
Points in Fig.~\ref{fig:compare_explore} refer to architectures sampled from both generators in the last 15 search steps where random explorers are disabled. 
The validation performances are color-coded.
We can see that our {\ourmodelshort} samples a set of graphs that have better validation accuracies while the ones of RNN generator have large variances in performances.
Moreover, the graphs in our case concentrate to those with smaller clustering coefficients, thus less likely being densely-connected.
On the contrary, RNN generator tends to sample graphs that are more likely to be densely-connected.
%
%
% {\bf RNN vs Graph Generator.}
While RNN has been widely used for NAS, we show in our experiments that our graph generator consistently outperforms RNN over three search spaces on two different datasets. 
This is likely due to the fact that our graph generator better leverages graph topologies, thus being more expressive in learning the distribution of graphs.

{\bf Bias in Evaluator.}
In our experiments, we use SuperNet evaluator, low-data, and full-data Oracle evaluator to efficiently evaluate the model. 
From the computational efficiency perspective, one would prefer the SuperNet evaluator.
However, it tends to give high rewards to those architectures used for training SuperNet.  
Although the low-data evaluator is more efficient than the full-data one, its reward is biased as discussed in Section \ref{sec:randwire_result}. 
This bias is caused by the discrepancy between the data distributions in low-data and full-data regimes. 
We also tried to follow \citep{tan2019mnasnet} to use early stopping to reduce the time cost of the full-data evaluator. 
However, we found that it assigns higher rewards to those shallow networks which converge much faster in the early stage of training.
% However, we observe consistent performance drop in Table \ref{tab:prob_randwire} and all methods tend to search for wide and shallow architectures. 
We show detailed results in Appendix \ref{app:early_stopping_bias}.

{\bf Search Space Design.}
The design of search space largely affects the performances of NAS methods. 
Our {\ourmodelshort} successfully learns good architectures on the challenging RandWire search space.
However, the search space is still limited as the cell graph across different stages is shared. 
A promising direction is to learn to generate graphs in a hierarchical fashion. 
For example, one can first generate a macro graph and then generate individual cell graphs (each cell is a node in the macro graph) conditioned on the macro graph.
This will significantly enrich the search space by including the macro graph and untying cell graphs.

{\bf Conclusion.}
In this paper, we propose a GNN-based graph generator for NAS, called {\ourmodelshort}. 
Our graph generator naturally captures topologies and dependencies of operations of well-performing neural architectures.
It can be learned efficiently through reinforcement learning. 
We extensively study its performances on the challenging RandWire as well as two widely used search spaces.
% With our graph generator, we are the first to explore the challenging randwire search space 
% We also did comprehensive experiments on enas-macro and nas-bench-101 where we demonstrate our proposed graph generator is consistently better.
Experimental results show that our {\ourmodelshort} consistently outperforms RNN-based generator on all datasets.
Future works include exploring ensemble methods based on our {\ourmodelshort} and hierarchical graph generation on even larger search spaces.

\subsubsection*{Acknowledgments}
This work was funded, in part, by the Vector Institute for AI, Canada CIFAR AI Chair, NSERC CRC, NSERC DG and Discovery Accelerator Grants, and Oracle Cloud credits.
Resources used in preparing this research were provided, in part, by the Province of Ontario, the Government of Canada through CIFAR, and companies sponsoring the Vector Institute \url{www.vectorinstitute.ai/#partners}, Advanced Research Computing at the University of British Columbia, and the Oracle for Research program. 
Additional hardware support was provided by John R. Evans Leaders Fund CFI grant and Compute Canada under the Resource Allocation Competition award.
We would like to thank Raquel Urtasun, Wenyuan Zeng, Yuwen Xiong, Ethan Fetaya, and Thomas Kipf for supporting the exploration along this research direction before this work.
% \newpage
\bibliography{main_ref}
\bibliographystyle{iclr/iclr_conference}

\newpage
\appendix

\section{More details on Graph Generator}
\begin{figure}[htbp]
    \centering
    \includegraphics[width=\textwidth]{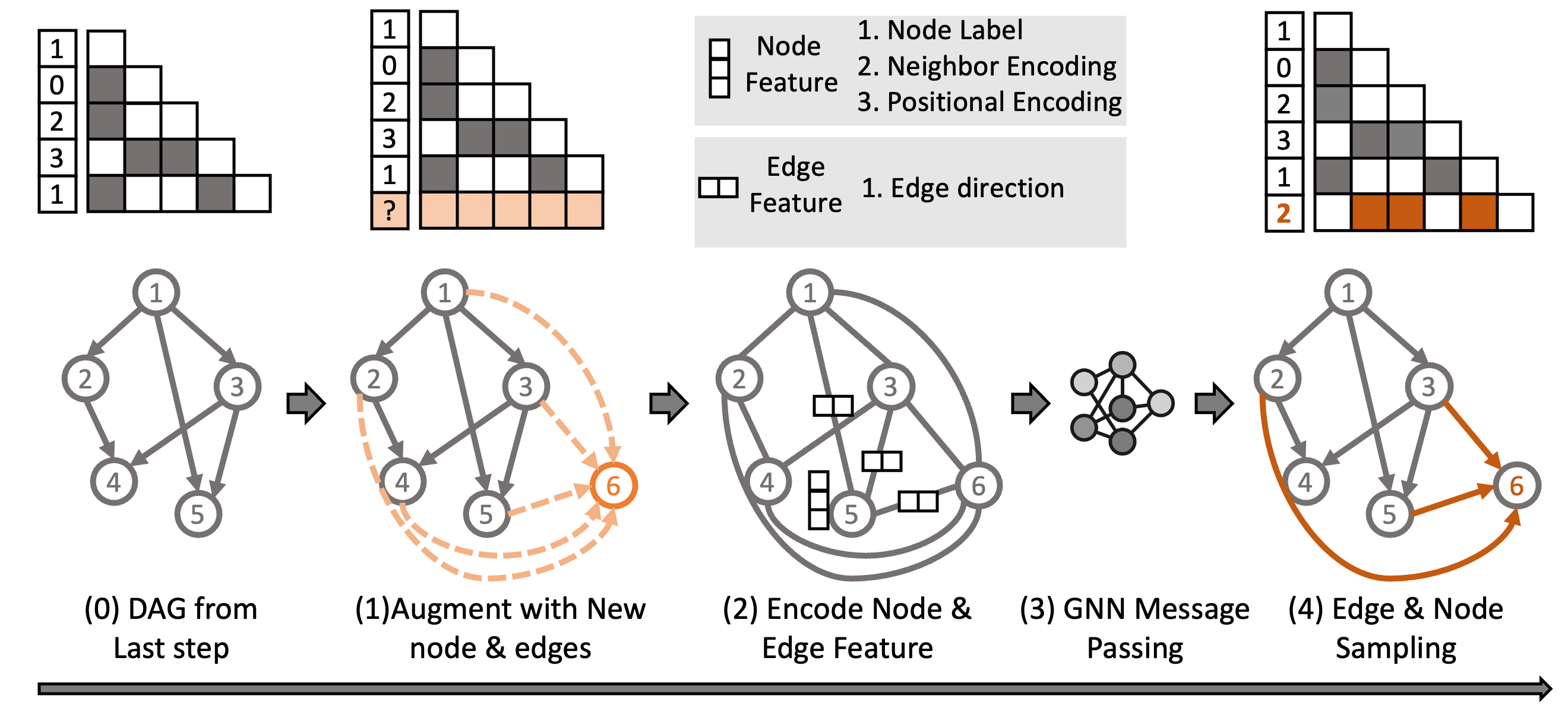}
    \caption{Detailed steps of auto-regressive generation with our graph generator.}
    \label{fig:graph_generation}
\end{figure}

To more clearly illustrate the sampling process of our generator, we detailed probabilistic sampling process of our generator in Fig. \ref{fig:graph_generation}.

% \color{black}

\section{Experiments on the NAS-Bench-201 Search Space}
\label{sec:nasbench_201}
% Introduce nasbench201
Here we compare our method on NAS-Bench-201\citep{dong2020bench} with random search (RANDOM)~\cite{bergstra2012random}, random search with parameter sharing (RSPS)~\cite{li2020random},  REA~\cite{real2019regularized}, REINFORCE~\cite{williams1992simple}, ENAS~\cite{pham2018efficient}, first order DARTS (DARTS 1st)~\cite{liu2018darts}, second order DARTS (DARTS 2nd), GDAS~\cite{dong2019searching}, SETN~\cite{dong2019one}, TAS~\cite{dong2019network}, FBNet-V2~\cite{wan2020fbnetv2}, TuNAS~\cite{bender2020can}, BOHB~\cite{falkner2018bohb}. 

\begin{table*}[htbp]
% \color{blue}
\centering
% \footnotesize
\setlength{\tabcolsep}{4pt}
\resizebox{\textwidth}{!}{
\begin{tabular}{| c | c | c | c | c | c | c |}
% \toprule
\hline
 \multirow{2}{*}{Methods} & \multicolumn{2}{|c}{CIFAR-10} & \multicolumn{2}{|c}{CIFAR-100} & \multicolumn{2}{|c|}{ImageNet-16-120} \\
\cline{2-7}
   & validation     & test           &   validation   &   test    & validation & test  \\
\hline
\multicolumn{7}{|l|}{\textit{Weight Sharing NAS:}}\\
\hline
  RSPS     & 87.60$\pm$0.61  & 91.05$\pm$0.66 & 68.27$\pm$0.72 & 68.26$\pm$0.96 & 39.73$\pm$0.34 & 40.69$\pm$0.36\\
  DARTS (1st) & 49.27$\pm$13.44 & 59.84$\pm$7.84 & 61.08$\pm$4.37 & 61.26$\pm$4.43 & 38.07$\pm$2.90 & 37.88$\pm$2.91 \\
  DARTS (2nd) & 58.78$\pm$13.44 & 65.38$\pm$7.84 & 59.48$\pm$5.13 & 60.49$\pm$4.95 & 37.56$\pm$7.10 & 36.79$\pm$7.59 \\
  GDAS     & 89.68$\pm$0.72  & 93.23$\pm$0.58 & 68.35$\pm$2.71 & 68.17$\pm$2.50 & 39.55$\pm$0.00 & 39.40$\pm$0.00 \\
  SETN     & 90.00$\pm$0.97  & 92.72$\pm$0.73 & 69.19$\pm$1.42 & 69.36$\pm$1.72 & 39.77$\pm$0.33 & 39.51$\pm$0.33 \\
  ENAS     & 90.20$\pm$0.00  & 93.76$\pm$0.00 & 70.21$\pm$0.71 & 70.67$\pm$0.62 & 40.78$\pm$0.00 & 41.44$\pm$0.00 \\
\hline
\hline
\multicolumn{7}{|l|}{\textit{Multi-trial NAS:}}\\
\hline
% \textit{} & & & & & &\\
  REA       & 91.25$\pm$0.31 & 94.02$\pm$0.31 & 72.28$\pm$0.95 & 72.23$\pm$0.84 & 45.71$\pm$0.77 & \textbf{45.77$\pm$0.80}\\
  REINFORCE & 91.12$\pm$0.25 & 93.90$\pm$0.26 & 71.80$\pm$0.94 & 71.86$\pm$0.89 & 45.37$\pm$0.74 & 45.64$\pm$0.78 \\
  RANDOM    & 91.07$\pm$0.26 & 93.86$\pm$0.23 & 71.46$\pm$0.97 & 71.55$\pm$0.97 & 45.03$\pm$0.91 & 45.28$\pm$0.97 \\
  BOHB      & 91.17$\pm$0.27 & 93.94$\pm$0.28 & 72.04$\pm$0.93 & 72.00$\pm$0.86 & \textbf{45.55$\pm$0.79} & 45.70$\pm$0.86 \\
% \midrule
\hline
  Ours      & \textbf{91.47$\pm$0.15} & \textbf{94.28$\pm$0.17} & \textbf{72.47$\pm$0.84} & \textbf{72.34$\pm$0.81} & 45.00$\pm$0.83 & 45.40$\pm$0.97\\
%   & Ours      & 91.17$\pm$0.27 & 93.94$\pm$0.28 & 72.04$\pm$0.93 & 72.00$\pm$0.86 & 45.55$\pm$0.79 & 45.70$\pm$0.86 \\
% \cmidrule{2-9}
%  & \multicolumn{2}{|c|}{\textit{ResNet}} & 90.86 & 93.91 & 70.50 & 70.89 & 44.10 & 44.23 \\
%   \multicolumn{2}{|c|}{\textbf{Optimal}} & 91.61 & 94.37 (94.37) & 73.49 & 73.51 (73.51) & 46.73 & 46.20 (47.31) \\
% \midrule
% \bottomrule
\hline
\end{tabular}
}
\caption{
% \color{blue}
Searched best architecture performance on NAS-Bench-201. We run our methods 10 times to obtain mean and standard deviation.
}
\label{table:benchmarking}
\end{table*}
To fairly compare with scores reported in \citep{dong2021nats}, we fix a search budget of 20000s on CIFAR10 and CIFAR100, and 30000s on ImangeNet-16-120, which is approximately 150, 80, and 40 oracle evaluations (with 1 sample evaluated per step) on CIFAR10, CIFAR100, and ImageNet-16-120 respectively. Specifically for NAS-Bench-201, we use random explorer in the first 10 steps and keep the top 15 architectures in the replay buffer. We found that our model outperforms previous methods on CIFAR10 and CIFAR100 datasets and is on par with state-of-the-art methods on ImageNet-16-120 dataset. \textbf{On ImageNet-16-120, after we extend the number of search budget from 40 to 60 steps, we significantly boost the performance to 45.57(+0.57) and 45.79(+0.4) on validation and test sets respectively.} This indicates that the search process of our model hasn't converged due to the limited search steps. This suggests that a reasonable number of search steps is needed for our model to reach its full potential.

\section{Comparison with NAGO}
Following \citet{xie2019exploring}'s work on random graph models, \cite{ru2020nago} propose to learn parameters of random graph models using bayesian optimization.
We compare with the randwire search space (refers to as RANG) in \citep{ru2020nago}. Since the original search space in \citep{xie2019exploring} do not reuse cell graphs for different stages, we train conditionally independent graph generators for different stages respectively. That is 3 conditionally independent generators for conv$_3$, conv$_4$, and conv$_5$ stage in Table \ref{tab:randwire_arch}. We perform a search on the CIFAR10 dataset, where each model is evaluated for 100 epochs. We restrict the search budget to 600 oracle evaluations. We align with settings in \citep{ru2020nago} for retraining and report sampled architecture's test accuracy and standard deviation in the  Table \ref{tab:nago_c10}.
\begin{table}[htbp]
    \centering
    \begin{tabular}{|c|c|c|c|}
    \hline
    Methods & Reference & Avg. Test Accuracy (\%) & Std. \\ 
    \hline
    RANG-D & \cite{xie2019exploring} & 94.1 & 0.16  \\
    RANG-BOHB & \cite{ru2020nago} & 94.0 & 0.26 \\
    RANG-MOBO & \cite{ru2020nago} & 94.3 & 0.13 \\
    Ours & -& \textbf{94.6} & 0.18\\
    \hline
    \end{tabular}
    \caption{Comparison of the searched results on CIFAR10. Mean test accuracy and the standard deviation are calculated over 8 samples from the searched generator. We align the search space design and retraining setting for a fair comparison.}
    \label{tab:nago_c10}
\end{table}
We can see that our method learns a distribution of graphs that outperforms previous methods.

\section{More details on the Randwire experiments}
\subsection{Details of RandWire search space}
\label{app:randwire}

\begin{figure}[htbp]
    \vspace{-0.1in}
    \centering
    \includegraphics[width=0.8\linewidth]{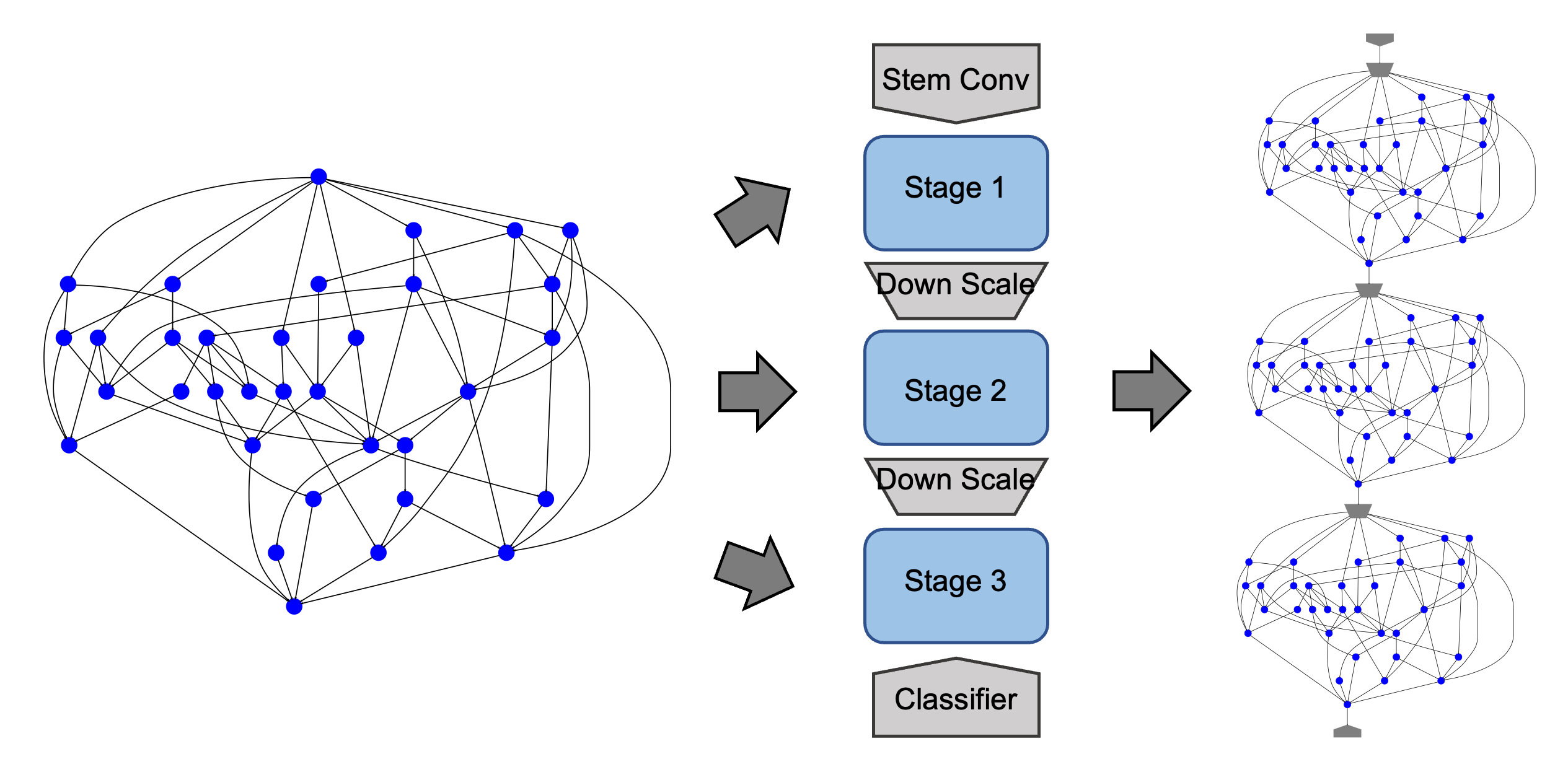}
    \caption{Visualization of RandWire search base space used in this paper. Different from \citep{xie2019exploring}, $\mathcal{G}$ here is shared across three stages.}
    \label{fig:randwire_arch}
    \vspace{-0.1in}
\end{figure}

Here we provided more details on the RandWire search space shown in  Table \ref{tab:randwire_arch} and Fig. \ref{fig:randwire_arch}.
%%%%%%%%%%%%%%%%%%%%%%%%%%%%%%%%%%%%%%%%%%%%%%%%%%%%%%%%%%%%%%%%%%%%%%%%%%%%%%%%%%%%%%%%%%%%%%%%%%%
\newcommand{\ft}[1]{\fontsize{#1pt}{1em}\selectfont}
\newcommand{\graph}{$\mathcal{G}$}
\newcommand{\m}{$\times$}
\begin{table}[htbp]
\begin{center}
\begin{tabular}{|c|c|c|c|c|c|}
\hline
 \multirow{2}{*}{Stage} & \multirow{2}{*}{Output} & \multicolumn{2}{c|}{{Base}} & \multicolumn{2}{c|}{{Large}}\\
    \cline{3-6}
  & & Cell & Channels& Cell& Channels\\
\hline
\multirow{1}{*}{$\text{conv}_1$} & \multirow{1}{*}{\ft{7} 112\m112}
&  {\ft{7} $\texttt{conv}_{3\times3}$ } & {32} &  {\ft{7} $\texttt{conv}_{3\times3}$} & {48}\\
\hline
$\text{conv}_2$ & {\ft{7} 56\m56}
  & {\ft{7} $\texttt{conv}_{3\times3}$} & 64 & {\ft{7} $\texttt{conv}_{3\times3}$} & 96\\
\hline
$\text{conv}_3$ & {\ft{7} 28\m28} & \graph & 64 & \graph & 192\\
\hline
\multirow{2}{*}{$\text{conv}_4$} & \multirow{2}{*}{\ft{7} 14\m14} & \multirow{2}{*}{\graph} & \multirow{2}{*}{128} & \graph & 288\\
\cline{5-6}
& & & & \graph &  384 \\
\hline
$\text{conv}_5$ & {\ft{7} 7\m7}
& \graph&256 & \graph & 586\\
\hline
\multirow{2}{*}{$\text{classifier}$} & \multirow{2}{*}{\ft{7} 1\m1}
& \multicolumn{4}{c|}{\ft{7} 1\m1 $\texttt{conv}_{1\times1}$, 1280-d} \\
& & \multicolumn{4}{c|}{\ft{7} global average pool, 200-d \emph{fc}, softmax} \\
\hline
\end{tabular}
\end{center}
\vspace{-.5em}
\caption{Randwire search space with base and large settings. Base is the default setting for search while Large refers to the architecture of scaled up models in Table \ref{tab:best_randwire}.
\texttt{conv} denote a ReLU-SepConv-BN triplet .
The input size is 224$\times$224 pixels. The change of the output size implies a stride of 2 (omitted in table) in the convolutions that are placed at the end of each block. $\mathcal{G}$ is the shared cell graph that has $N=32$ node.
}
\label{tab:randwire_arch}
\end{table}

%%%%%%%%%%%%%%%%%%%%%%%%%%%%%%%%%%%%%%%%%%%%%%%%%%%%%%%%%%%%%%%%%%%%%%%%%%%%%%%%%%%%%%%%%%%%%%%%%%%
\subsection{Details for RandWire experiments}
For experiment on Tiny-Imagenet, we resize image to $224\times 224$ as showed in Table \ref{tab:randwire_arch}. We apply the basic data augmentation of horizontal random flip and random cropping with padding size 4.
We provide detailed hyper-parameters for oracle evaluator training  and learning for {\ourmodelshort} in Table \ref{tab:hyper_parameter_randwire}
\begin{table}[htbp]
    \centering
    \begin{tabular}{|l|l|l|l|}
        \hline
        \multicolumn{2}{|c|}{Oracle Evaluator}& \multicolumn{2}{c|}{Graph Controller}\\
        \hline
        batch size & 256& graph batch size & 16\\
        optimizer & SGD & generator optimizer  & Adam\\
        learning rate & 0.1& generator Learning rate& 1e-4\\
        learning rate deacy & consine lr decay & generator learning rate decay & none\\
        weight decay & 1e-4 & generator weight decay & 0.\\
        grad clip& 0. & generator gradient clip & 1.0\\
        training epochs & 300 & replay buffer fitting epochs& 2000\\
        \hline
    \end{tabular}
    \caption{Hyperparameter setting for oracle evaluator and training our graph generator.}
    \label{tab:hyper_parameter_randwire}
\end{table}

\subsection{Visualization of architectures from our generator}
\begin{figure}[htbp]
% \centerfloat
     \includegraphics[width=\textwidth]{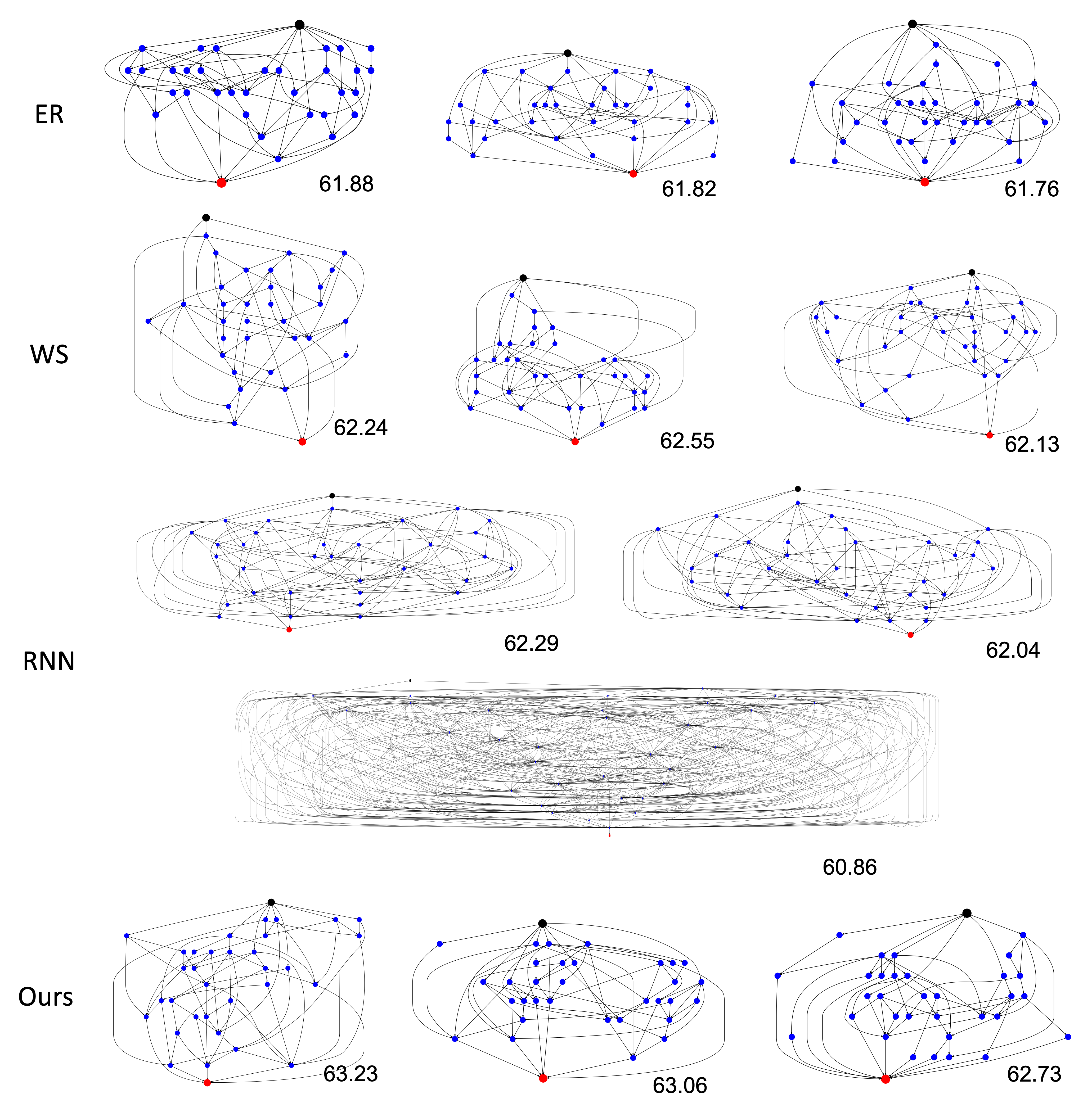}
    \caption{Visualization of Top 3 architectures sampled by each method. We observe that around 50\% of samples from RNN generators are densely connected graphs or even fully connected graphs.}
    \label{fig:vis_randwire}
\end{figure}

Here we visualize the top candidate architectures in Fig. \ref{fig:vis_randwire}

\subsection{Bias for early stopping}
\label{app:early_stopping_bias}
As discussed in Section~\ref{sec:discuss}, using early stopping will lead to local minimal where the generator learns to generate shallow cell structure. We quantify this phenomenon in table \ref{tab:ablate_oracle}, where we can see that with early stopping training, the generator will generate more shallow architectures with a shorter path from input to output. The corresponding average final validation accuracy also dropped by a large margin compared to the low data evaluator counter part.
\begin{table}[htbp]
    \centering
    \begin{tabular}{|c|c|c|c|}
    \hline
        evaluator& Final Val Acc& Average Path & Longest Path\\
        \hline
        early stopping& 61.86 & 2.595 & 6.125\\
        \hline
        low data regime & 62.57 & 3.046 & 8.75\\
        \hline
    \end{tabular}
    \caption{In the table, we show ablation on the choice of oracle evaluator with our graph generator. The average Path and Longest path are computed as the average path length and longest path length from input to output over 8 samples from the corresponding generator.}
    \label{tab:ablate_oracle}
\end{table}

\section{More details on ENAS Macro experiments}
\label{app:enas}
For ENAS Macro search space, we use a pytorch-based open source implementation\footnote{https://github.com/microsoft/nni/tree/v1.6/examples/nas/enas} and follow the detailed parameters provided in \citep{pham2018efficient} for RNN generator.
Specifically, we follow \citep{pham2018efficient} to train the SuperNet and update the generator in an iterative style.
At each search step, two sets of samples $\mathcal{G}_{\text{train}}$ and $\mathcal{G}_{\text{eval}}$ are sampled from the generator. 
$\mathcal{G}_{\text{train}}$ is used to learn the SuperNet's weights by back-propagating the training loss. 
The updated SuperNet is used for evaluating $\mathcal{G}_{\text{eval}}$, which is then used for updating the generator. 

For our generator, we evaluate 100 architectures per step and update our generator every 5 epochs of SuperNet training. Instead of evaluating on a single batch, we reduce the number of models evaluated per step and evaluate on the full test set. We found this stables the training of our generator while keeping evaluation costs the same. In the replay buffer, the top 20\% of architectures is kept.

For training SuperNet and RNN generator, we follow the same hyper-parameter setting in \citep{pham2018efficient} except the learning rate decay policy is changed to cosine learning rate decay. For training our generator we use the same hyperparameter as in Table \ref{tab:hyper_parameter_randwire} with graph batch size changed to 32. For retraining the found the best architecture, we use a budget of 600 epoch training with a learning rate of 0.1, batch size 256, and weight decay 2e-4. We also apply a cutout with a probability of 0.4 to all the models when retraining the model.

\subsection{Visualization of best architecture found}
\begin{figure}
    \centering
    \includegraphics[width=\linewidth]{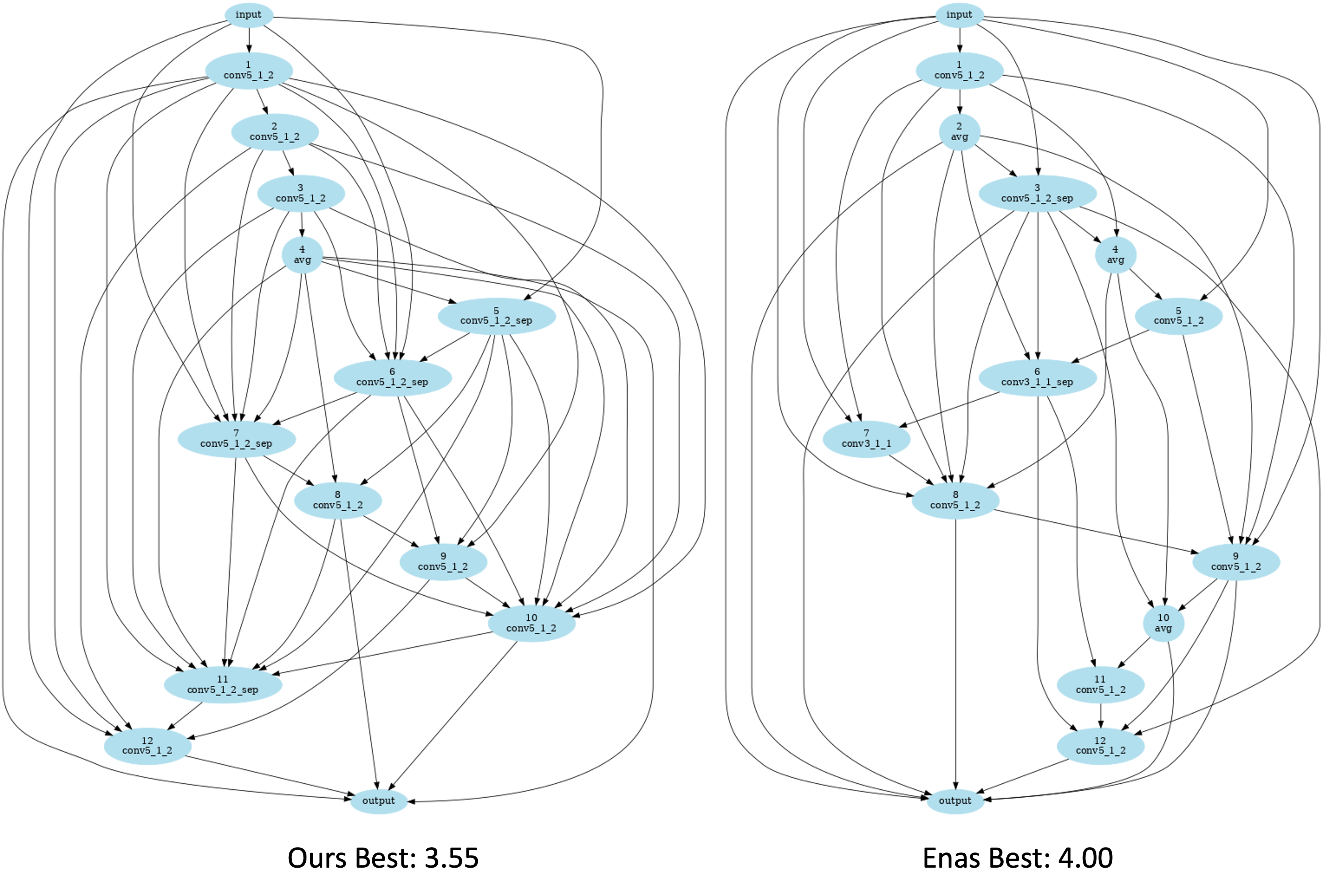}
    \caption{The best architecture found by {\ourmodelshort} and RNN generator \citep{pham2018efficient}. Correspond to scores report in table \ref{tab:enas_results}. To get this architecture, we  pre-evaluate 8 samples for both methods  and select top-performing architecture.}
    \label{fig:vis_enas}
\end{figure}

Here we visualize the best architecture found by {\ourmodelshort} and RNN generator for Enas Macro search space in Fig. \ref{fig:vis_enas}.

\section{More details on NAS-Bench-101}
For sampling on NAS-Bench-101 \citep{ying2019bench}, we first sample a 7-node DAG, then we remove any node that is not connected to the input or the output node. We reject samples that have more than 9 edges or don't have a path from input to output.

To train our generator on Nas-Bench-101, we use Erdős–Rényi with $p=0.25$, $\epsilon$ is set to 1 in the beginning and decreased to 0 after 30 search steps.

For the replay buffer, we keep the top 30 architectures.
Our model is updated every 10 model evaluations, where we train 70 epochs on the replay buffer at each update time. The learning rate is set to 1e-3 for with a batch size of 2 on Nas-bench-101.
\end{document}